\documentclass[runningheads]{llncs}
\usepackage[T1]{fontenc}
\usepackage{graphicx}
\usepackage{multirow}
\usepackage{caption}
\begin{document}
\title{Event Detection for Active Lower Limb Prosthesis}
%
%
\author{J. D. Clark\inst{1}\orcidID{0009-0007-7423-647X}, P Ellison\inst{1}\orcidID{0000-0003-3741-8219}}
\authorrunning{J. D. Clark and P. Ellison}
%
\institute{University of York, York YO10 5DD, United Kingdom} 
\maketitle              
\begin{abstract}
Accurate event detection is key to the successful design of semi-passive and powered prosthetics. Kinematically, the natural knee is complex, with translation and rotation components that have a substantial impact on gait characteristics. When simplified to a pin joint, some of this behaviour is lost. This study investigates the role of cruciate ligament stretch in event detection. A bicondylar knee design was used, constrained by analogues of the anterior and posterior cruciate ligaments. This offers the ability to characterize knee kinematics by the stretch of the ligaments. The ligament stretch was recorded using LVDTs parallel to the ligaments of the Russell knee on a bent knee crutch. Which was used to capture data on a treadmill at 3 speeds. This study finds speed dependence within the stretch of the cruciate ligaments, prominently around 5\% and 80\% of the gait cycle for the posterior and anterior. The cycle profile remains consistent with speed; therefore, other static events such as the turning point feature at around 90\% and 95\% of the cycle, for the posterior and anterior, respectively, could be used as a predictive precursor for initial contact. Likewise at 90\% and 95\%, another pair of turning points that in this case could be used to predict foot flat. This concludes that the use of a bicondylar knee design could improve the detection of events during the gait cycle, and therefore could increase the accuracy of subsequent controllers for powered prosthetics. 

\end{abstract}
\section{Introduction}
Most people will not give a thought to the complex intricacy that enables us to walk, almost subconsciously, every day. However, as you can imagine it is made very clear to an amputee. Prosthesis, literally meaning "a placing addition", in any form, primarily has one function: to add what is "missing". In the context of limb prosthesis, there are two main branches of development, passive and powered. Passive prosthetics provide the form and function of a limb, without the ability to input energy into the system. They can offer energy storage, but they rely on the user to compensate for the lack of energy with their residual limb, causing asymmetries \cite{hobara_lower_2011}. Powered prosthetics aim to bridge the power gap, and introduce energy into the system as the missing muscles would. They are, however, not without their drawbacks. Often designed from classical robotics principles first, the joint structures are either a bar linkage or a pin-jointed mechanism. These kinematics do not match the behaviour of human biological joints, simplifying the complexity of the joint motion. This causes usability issues to a certain degree, resulting in an extended learning curve for use, as the gait now differs extensively. As the kinematics and kinetics of the joint now differ, there is a hard limit to the optimization of event detection in a prosthesis due to the loss of detail that the matching neurological system would have. Using a joint that matches the motion of the human knee will allow for a more typical gait and provide more information for a control system, theoretically resulting in a more natural gait if implemented. The control of current active prosthetics is based heavily on finite state machines or pattern recognition to function \cite{song_continuous_2024}. These control algorithms are simply replaying predetermined patterns with little to no direct input from the user \cite{song_continuous_2024}. Sensors in the device are crucial for feedback into the control system but are implemented as auxiliary type checks for state progression, not direct control \cite{song_continuous_2024,ficanha_control_2015}. Essentially, the adaptability of current controllers is limited to the conditions that have been identified and instituted offline \cite{song_continuous_2024}.\\\\   
The detection of speed in a prosthesis will differ with design, but fundamentally, an encoder can be used on the axis as an information stream that can be used to ascertain the speed of the prosthesis from the joint angle. Rotary encoders will only work with a fixed rotational axis, such as a pin joint; therefore the joint angle estimation will again suffer from the simplification of the biomechanics. As the human knee has a convex and concave joint interface, the joint has a translational component as well as its rotational component; this excludes the use of a rotary encoder, as it would cancel out the added information of the joint. A direct physiological measure, rather than the abstracted joint angle, such as the ligament displacement of the anterior and posterior cruciate ligaments (ACL, PCL) of the knee, would provide the data stream needed for the control system. The Russell knee joint is a biomimetic human knee joint that is capable of replicating human-like motion, restricted to the sagittal plane \cite{russell_biomimicking_2018,russell_kinematic_2018}. Like the human knee joint, it uses mechanical surrogates of the ACL and PCL to constrain the motion of the joint, where the displacement of these ligaments where used as a design constraint to produce the joint surfaces used \cite{russell_biomimicking_2018,russell_kinematic_2018}. Therefore, using this design, we can characterize motion with two independent sources, ligament stretch, where previously we had only one, knee angle. Where knee angle has previously been used as an auxiliary check in current controllers, this added and more realistic detail could provide the opportunity for sensor data to provide a more active role in the control architecture \cite{song_continuous_2024,ficanha_control_2015}. Potentially allowing for the online detection of key events in the cycle, such as initial contact and toe-off, which previously were a predetermined state, not an online prediction\cite{song_continuous_2024,ficanha_control_2015}.    

\newpage
\section{Materials and Methods}
\subsection{Design of Bend Knee Rig}

\begin{figure}[h]
\centering
\begin{minipage}{.5\textwidth}
  \centering
  \includegraphics[width=.55\linewidth]{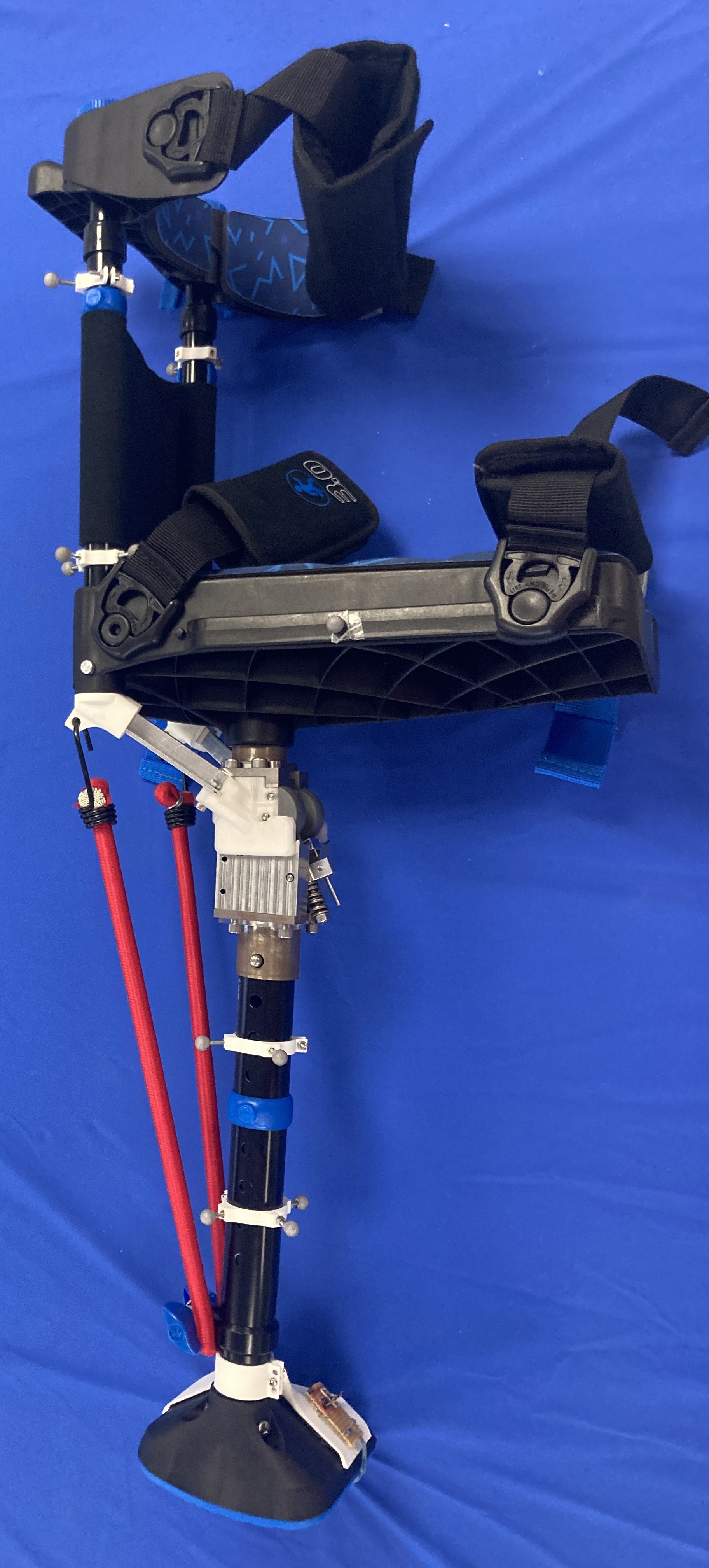}
  \captionsetup{width=.\textwidth}
  \caption{The iWalk 3.0 bent knee crutch, fitted with the Russell knee for experimentation with a "healthy" volunteer.}
  \label{rig}
\end{minipage}%
\begin{minipage}{.5\textwidth}
  \centering
  \includegraphics[width=.75\linewidth]{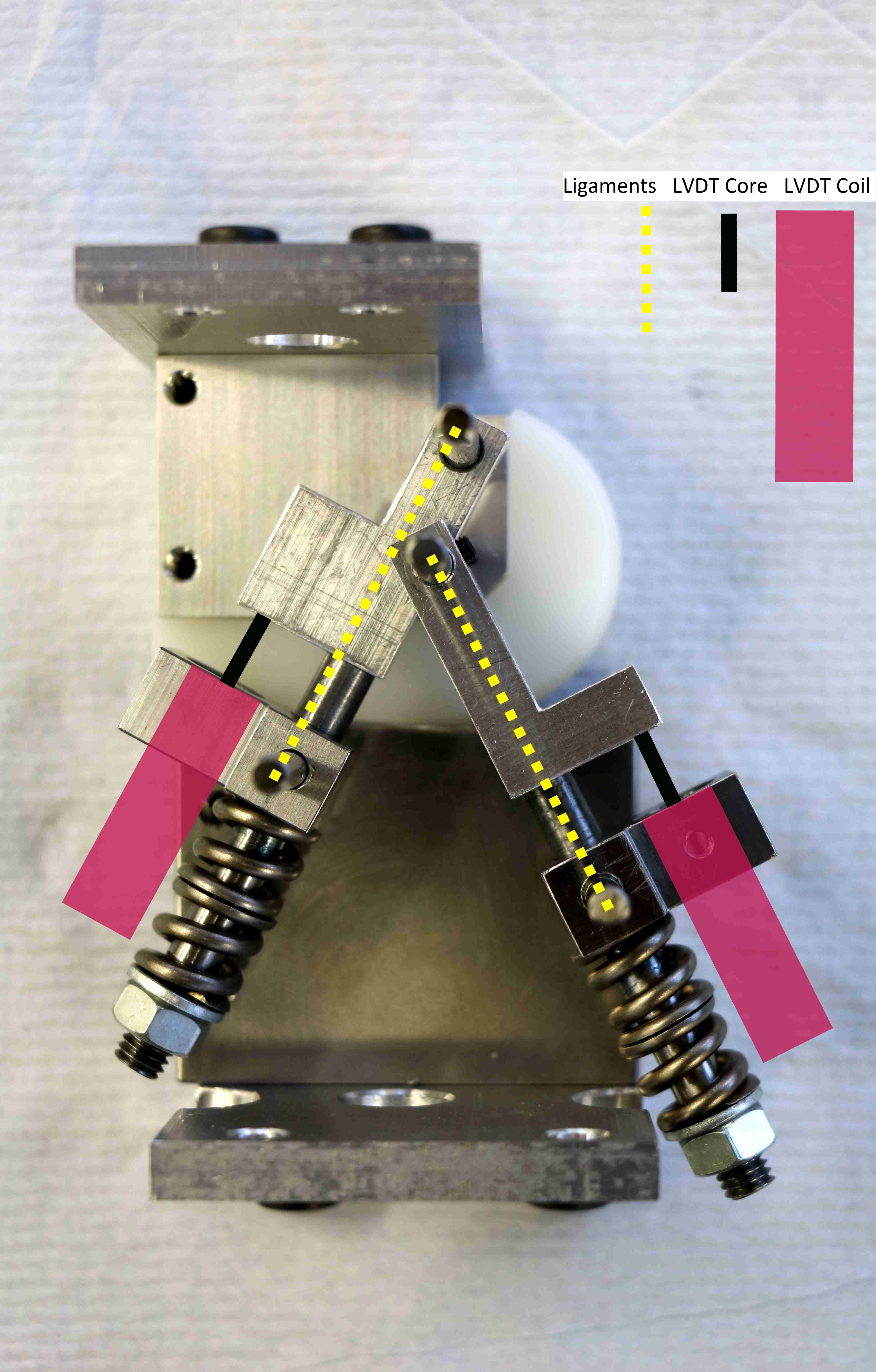}
  \captionsetup{width=.75\textwidth}
  \caption{A cross-section of the Russell knee with the LVDT placements shown. Adapted from \cite{russell_complete_2020}}
  \label{knee}
\end{minipage}
\end{figure}

\noindent Fig.~\ref{rig} shows the prototype prosthesis of the Russell knee, Fig.~\ref{knee}, bicondylar articulating geometry, and elastic springs that represent the cruciate ligaments. The knee joint was fitted to the bent knee crutch (iWalk 3.0 iWALKFree, US) to allow healthy subjects to operate the prosthesis. Linear variable different transformers (LVDT) (Solartron SM3, UK) were fitted parallel to the springs allowing measurement of stretch during joint movement. The movement of the joint is passive; therefore, to counteract the cantilever effect from the body, an elastic cord was used with an approximate stiffness, allowing the device to be straight when unlocked in the upright position. Finally, a force-sensitive resistor (FSR) was mounted to the rear of the foot of the I-Walk 3.0, so that initial contact could be measured. Two LVDT signal conditioner units were mounted in their own box, and the outputs of which were fed into a DAQ card (6211 National Instruments, US). The voltage output from the potential divider for the FSR was also fed to another input on the DAQ card. A treadmill was then placed in the centre of the capture space, and the LVDT cables were extended to accommodate a safe distance from the user and the treadmill. All data was recorded directly from the DAQ card into MATLAB for processing

\subsection{Analysis and Testing}
The procedure set for testing is as follows:
\begin{enumerate}
    \item The volunteer will walk three times for 1 minute to allow for a minimum of 10 consistent strides to be collected at three speeds. 1.0 kph, 1.5 kph, and 2.0 kph.
    \item All data will be simultaneously collected and organised by speed and volunteer.
\end{enumerate}
The voltage-time series of the LVDTs is converted to a displacement-time series separately. As each LVDT had its own calibrated sensitivity, the scalar conversion is bespoke to each ligament. For all sources individually, the average of at least 10 cycles is now calculated by assigning a common time grid, which is chosen as the longest cycle. Each cycle was then interpolated to match the length of the common time grid and then averaged to achieve a mean cycle for each source. The standard deviation of each trial was also calculated, allowing for a test of equal variance. As the variance was assumed equal, a one-tailed t-test was used to check if points of interest had a significant difference in speeds. The criteria for points of interest were consistent patterns at points of interest, such as initial contact, toe off, and gait phase transition \cite{vu_review_2020}. As well as arbitrary points with a large variance between speeds that could be used to detect speed.      
\section{Results}
\begin{figure}[h]
    \centering
    \includegraphics[width=0.75\linewidth]{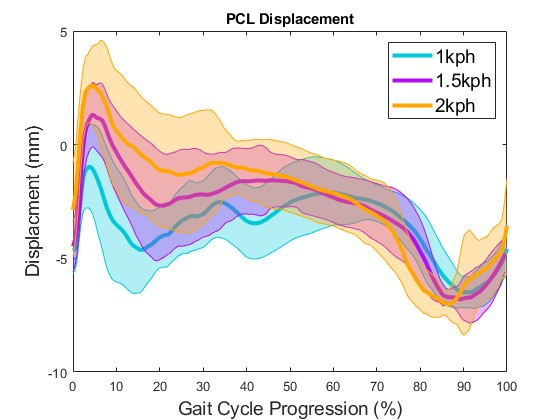}
    \caption{Averaged ligament displacement for the PCL over the gait cycle at 3 speeds. The shaded regions indicate the ± 1 standard deviation.}
    \label{pcl}
\end{figure}
\begin{figure}[h]
    \centering
    \includegraphics[width=0.75\linewidth]{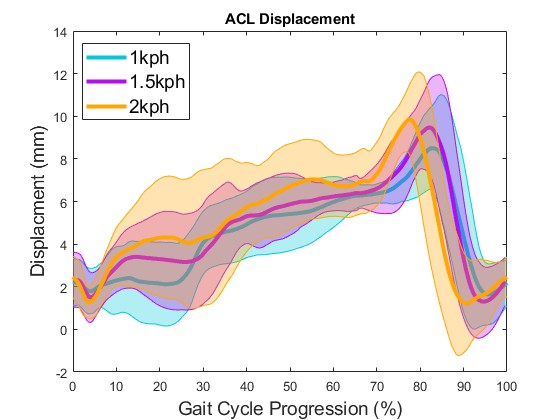}
    \caption{Averaged ligament displacement for the ACL over the gait cycle at 3 speeds. The shaded regions indicate the ± 1 standard deviation.}
    \label{acl}
\end{figure}

\newpage
\subsection{Key Events}
Separate from speed, the general shape of the profile has some interesting features. Most interestingly there is a similar effect at 80-90\% in both the PCL and ACL, the local minimum, where the ligaments are relaxing before the next cycle. We also see a similar, but less prominent, effect at 0-5\% of the cycle, however, with opposing directions of trend and less similar amplitudes.   
\subsection{Speed}
\noindent While the standard deviation is shown as the bounds in Fig.~\ref{pcl} and Fig.~\ref{acl}, there is a clear trend in the graph due to a variation in speed. While there is evidence here of multiple speed-dependent areas, most prominent are the peaks and around 80\% of the gait cycle in the ACL and around 5\% percent of the gait cycle for the PCL. The difference in the means between the 1.5 kph and 2.0 kph speeds compared to the 1.0 kph is shown in Tab.\ref{tab1}.

\begin{table}[h]
\centering
\caption{Table of results for the identified speed-dependent peaks of the ligament displacement.}
\label{tab1}
\begin{tabular}{|l|l|l|l|l|l|l|l|l|}
  \hline
  \multirow{2}{*}{Event} &
    \multicolumn{2}{c|}{1.0 kph (n=27)} &
    \multicolumn{3}{c|}{1.5 kph (n=31)} &
    \multicolumn{3}{c|}{2.0 kph (n=18)} \\
  \cline{2-9}
  & $\bar{x}$ & $\sqrt{\sigma}$ & $\bar{x}$ & $\sqrt{\sigma}$ & $p$ & $\bar{x}$ & $\sqrt{\sigma}$ & $p$ \\
  \hline
  80\% of PCL & 8.499 & 2.002 & 9.472 & 2.162 & 0.041 & 9.853 & 1.677 & 0.012 \\
  \hline
  5\% of ACL & -0.962 & 1.892 & 1.311 & 1.414 & $\approx 0$ & 2.601 & 1.699 & $\approx 0$ \\
  \hline
\end{tabular}
\end{table}

\section{Discussion}
The local minima at 80-90\% of the cycle for both ACL and PCL could potentially be a powerful observation, as they seem to occur consistently before initial contact. This would indicate that it is roughly the start of the terminal swing, and therefore a consistent predictive indicator for initial contact \cite{vu_review_2020}. With this, an algorithmic prediction could be made before determining the initial contact time rather than a predetermined state change with a loop delay to account for variations \cite{song_continuous_2024,ficanha_control_2015}. This could also leave enough time for the control to prepare the prosthesis for the event of the current cycle, not a predetermined profile \cite{song_continuous_2024}. Although the amplitude is weaker on the ACL, the maxima and minima, and 0-10\% of the cycle would align with the loading response and therefore could be used similarly to predict foot flat \cite{vu_review_2020}. \\\\ 
As for speed, as expected, the trends show that this occurs at the maximum extension of the ligament. This result is consistent with the expectation that knee angle increases with speed. As a result, the data indicates the ligament stretch is dependent on the gait speed as it is directly related to knee angle. While the inter-step variation of gait is large, the average with a large sample size provides evidence that information about gait speed can be extracted from the displacement. The peaks are the more obvious areas, but due to the variance, sensor fusion would likely have to be employed to give a per-cycle estimate. Estimates combined from both ligaments would reduce uncertainty, but when coupled with electromyography of the local muscles, a robust estimation of gait speed could be brought closer to a per cycle basis \cite{vu_review_2020}. Also of note is that during the cycle, there appears to be a convergence of speed, or speed-independent areas. These appear to lead to key events, such as 60\% of the transition from the stance to the swing phase \cite{vu_review_2020}. This furthers the key events, or events at the "zero" speed, showing the consistency of the waveform independent of speed.

\section{Conclusion}
In conclusion, this study has shown that the use of analogue ligaments in a knee prosthesis could aid to significantly improve current event detection methods within active lower limb prosthetic devices. While the complexity of the joint increases, it is justified by the gain of two data streams with an increase in kinematic data. If taken advantage of, it could robustly predict key high-energy events such as initial contact during the cycle. This could be driven by the ligament data, where previously this would have been predetermined using a finite state machine, which uses knee angle as an auxiliary check  \cite{ficanha_control_2015,song_continuous_2024,vu_review_2020}. It has also been shown that speed prediction could potentially be derived from the ligament stretch, allowing the system to make estimates of speed during the cycle from analogue physiological signals, rather than an implied measure such as knee angle.   

\bibliographystyle{splncs04}
\noindent
\bibliography{Talos_lib}

\end{document}